\documentclass[conference]{IEEEtran}
\IEEEoverridecommandlockouts
\usepackage{cite}
\usepackage{amsmath,amssymb,amsfonts,mathrsfs,amsthm}
\usepackage{algorithm}
\usepackage{algpseudocode}
\usepackage{graphicx}
\usepackage{textcomp}
\usepackage{xcolor}
\usepackage{url}
\usepackage{adjustbox}
\usepackage{balance}
\usepackage{makecell}

\begin{document}

\title{Three-Way Emotion Classification of EEG-based Signals using Machine Learning}

\author{\IEEEauthorblockN{Ashna Purwar, Gaurav Simkar, Madhumita, and Sachin Kadam}
\IEEEauthorblockA{{Electronics and Communication Engineering Department} \\
{Motilal Nehru National Institute of Technology (MNNIT) Allahabad, Prayagraj, UP 211004, India}\\
Email: \{ashna.20224038, gaurav.20224068, madhumita.20224091, sachink\} @mnnit.ac.in}
}

\maketitle

\begin{abstract}
Electroencephalography (EEG) is a widely used technique for measuring brain activity. EEG-based signals can reveal a person's emotional state, as they directly reflect activity in different brain regions. Emotion-aware systems and EEG-based emotion recognition are a growing research area. This paper presents how machine learning (ML) models categorize a limited dataset of EEG signals into three different classes, namely Negative, Neutral, or Positive. It also presents the complete workflow, including data preprocessing and comparison of ML models. To understand which ML classification model works best for this kind of problem, we train and test the following three commonly used models: logistic regression (LR), support vector machine (SVM), and random forest (RF). The performance of each is evaluated with respect to accuracy and F1-score. The results indicate that ML models can be effectively utilized for three-way emotion classification of EEG signals. Among the three ML models trained on the available dataset, the RF model gave the best results. Its higher accuracy and F1-score suggest that it is able to capture the emotional patterns more accurately and effectively than the other two models. The RF model also outperformed the existing state-of-the-art classification models in terms of the accuracy parameter.
\end{abstract}

\section{Introduction} \label{Sec:Intro}
Everything we think, feel, and do is controlled by the incredibly complex human brain. Electroencephalography, or simply EEG, is one of the most useful and noninvasive methods of studying its electrical activity. EEG records tiny voltage changes that happen when neurons communicate, and because it captures these signals with high temporal resolution and accuracy, it is useful as a tool in understanding various cognitive and emotional phenomena processes. Because EEGs can reveal emotional changes that are not readily apparent from facial expression or behavior, researchers in the fields of affective neuroscience and machine learning have also shown an increasing interest in using them over the past decade~\cite{li2021eeg,Li2022EEGEmotion,ding2022tsception,xu2023amdet,zhang2022ganser,houssein2022human,lim2024review}.

Identifying emotions from EEG data has numerous applications~\cite{erat2024emotion}. Emotion-aware computing systems enable personalized experiences in e-learning~\cite{lin2018mental}, gaming~\cite{vasiljevic2020brain}, and entertainment~\cite{de2023research}. EEG signal analysis in mental health can detect early signs of stress, anxiety, and depression~\cite{al2023classification,minkowski2021feature,tahira2023eeg}. Emotion detection is crucial for next-generation brain-computer interface (BCI) technologies, enabling real-time responses to users' mental and emotional states~\cite{yin2025eeg}.

However, dealing with EEG signals is not easy. The data is highly variable, noisy, and sensitive to even minor changes or environmental factors. Emotional reactions vary across individuals, complicating categorization. Hence, emotion classification models require proper preprocessing and feature selection for consistent performance. Recent research focuses on deep learning models to automatically learn feature patterns from raw data~\cite{jafari2023emotion,wang2023deep,geng2025deep}, while traditional methods rely on manually extracted features such as statistical measures~\cite{pillalamarri2025review} and frequency-band energies~\cite{wagh2022performance}.
Finding an emotion classification model that balances speed, accuracy, and interpretability remains a challenging task. Previous studies have shown promising results for classical machine learning (ML) techniques like support vector machine (SVM), random forest, and naive bayes~\cite{houssein2022human}. However, performance relies heavily on data preprocessing and feature selection. Using the same preprocessing pipeline across multiple models helps analyze algorithm performance and identify areas for improvement. 

The motivation for selecting this problem is the practical need for models that can handle smaller datasets. Collecting EEG data can be expensive and time-consuming, making deep learning models with large datasets not always feasible. The ML algorithms can function on smaller datasets, making them ideal for real-world applications with limited data.
In this paper, we use supervised ML techniques to classify EEG signals into three general emotional categories, namely Negative, Neutral, and Positive. Our goal is to evaluate the effectiveness of ML algorithms and comprehend their relative benefits and drawbacks when using EEG data. EEG-based signal preprocessing, feature extraction, model training, and evaluation are some of the steps involved in the operations. In this sense, this work intends to add to the body of knowledge regarding ML techniques for EEG-based emotion recognition.

The contributions of this paper are as follows:
\begin{itemize}
    \item EEG signal preprocessing and visualization for three-way emotion classification.
    \item  Study the impact of three ML algorithms, namely logistic regression (LR), SVM, and random forest (RF), on the classification of emotions using EEG signals.
    \item Comparison of the ML models' performances using metrics like accuracy and F1-score. Also, a comparison of the best-performing ML model with state-of-the-art models in the literature for the given limited EEG dataset. 
\end{itemize}



\section{Related Work} \label{Sec:RelatedWork}
The development of EEG-based emotion classification systems has been driven by a number of previous studies and open-source resources~\cite{jafari2023emotion,wang2023deep,geng2025deep,pillalamarri2025review,wagh2022performance,2022Vidhi,song2022eeg,acharya2020multi}. The creation and validation of ML models for EEG data has been made simpler with the help of numerous online practical workflows. For example, the GitHub repositories~\cite{2022Vidhi,song2022eeg,acharya2020multi} provide complete pipelines that include preprocessing, feature extraction, and classifier testing. These studies emphasize how important it is to filter noise, choose features, and adjust model parameters in order to create dependable EEG classification systems.
Peer-reviewed research has established a theoretical framework for this problem. Fernandes et al.~\cite{fernandes2024emotion} propose advanced neural architectures capable of recognizing complex patterns in EEG signals. Bazgir et al.~\cite{bazgir2018emotion} provide a summary of biosignal-based emotion recognition techniques, addressing common issues such as noise, subject variability, and feature engineering. Cheng et al.~\cite{cheng2022eeg}, Li et al.~\cite{li2018cross}, Alhagry et al.~\cite{alhagry2017emotion}, Zang et al.~\cite{zhang2020investigation}, and Zheng et al.~\cite{zheng2015investigating} demonstrate that CNNs, LSTMs, and hybrid architectures can learn nonlinear time-dependent emotional patterns from EEG data. Furthermore, they highlight potential issues such as overfitting and difficulty making generalizations about different people. Standardized scientific toolkits and documentation have greatly aided EEG research. Gramfort et al.~\cite{gramfort2014mne} state that MNE-Python provides precise techniques for EEG preprocessing, including signal filtering, artifact removal, data segmentation, and spectral analysis.

The motivation for this research stems from scientific curiosity and potential real-world implications. Creating a reliable emotion categorization system using EEG signals can result in more human-centered and compassionate technologies. Analyzing ML models will help us advance emotional computing, mental health care, adaptable interfaces, and intelligent decision-making systems.

\section{Background and Dataset}\label{Sec:Background_Dataset}
In this work, we build, train, and compare three supervised ML models, namely, logistic regression (LR), support vector machine (SVM), and random forest (RF). These models are chosen because they have a different approach toward learning and classification. Studying them together helps us to understand how different modelling strategies deal with the challenges of EEG-based emotion classification. Since EEG data is naturally noisy, high-dimensional, and time-varying, the choice of having appropriate models and evaluating them correctly is an important task. 

\subsection{Machine Learning Models}
In the following subsections, we provide a brief background of the three ML models.
\subsubsection{Logistic Regression} \label{Sec:LogisticRegression}
It is one of the most widely used algorithms for binary and multi-classification. The logistic (sigmoid) function $\sigma(x)$ is used to assess the probability that a data point $x$ belongs to a specific class rather than forecasting a continuous value as shown in~\eqref{eq:sigmod}.
\begin{equation}
    \label{eq:sigmod}
    \sigma(x) = \frac{1}{1 + e^{-x}}
\end{equation}
Because of its simplicity, it is commonly employed as a baseline model in a variety of classification problems, including EEG-based emotion recognition~\cite{guerrero2021eeg}. When applied to EEG data to measure different variables such as statistical measures, entropy values, or band-power data, the model identifies interesting patterns as long as the decision boundaries are not overly complex. However, there are practical limitations. Because of the detailed nature of brain signals, EEG data is typically nonlinear and complex. The LR model assumes log-odds linearity, making it less suitable for simulating extremely nonlinear interactions. Similarly, the LR model may struggle to deal with the high-dimensional or correlated information found in EEG datasets. To address this, regularization techniques (L1 or L2) are commonly used to prevent overfitting. L2 regularization is used in this study to stabilize the model and address feature correlations. Following training, the classifier's predictions are evaluated against standard performance criteria to determine its ability to categorize emotional states. The findings provide a solid foundation for understanding how linear discriminative models compare to more sophisticated techniques such as SVMs, RFs, and deep learning architectures.

\subsubsection{Support Vector Machine (SVM)} \label{Sec:SupportVectorMachine}
Support Vector Machines are a popular supervised learning technique. They detect a dividing line, or hyperplane, between different classes in a feature space~\cite{cortes1995support}.  The SVM classifier assigns a label 
\begin{equation}
    \label{eq:SVM}
    f(\mathbf{x}) =
    \sum_{i=1}^{N} \alpha_i \, y_i \, K(\mathbf{x}_i, \mathbf{x}) + b
\end{equation}
based solely on support vectors. In~\eqref{eq:SVM}, $\mathbf{x}$ denotes the test sample and $\mathbf{x}_i, i=1, \ldots, N,$ represents the support vector. The class labels are determined by $y_i \in \{-1,+1\}$. The coefficients $\alpha_i$ are the learned Lagrange multipliers, $K(\mathbf{x}_i, \mathbf{x})$ is the kernel function that maps the data into a higher-dimensional feature space, and $b$ is the bias term. Together, these components define the classifier function, $f(\mathbf{x})$, which assigns the predicted class according to $\hat{y} = \operatorname{sgn}(f(\mathbf{x}))$.\footnote{$\operatorname{sgn}(x)=(-1)\mathbf{1}_{x<0}+(1)\mathbf{1}_{x\ge0}$
}

The SVM classifiers are useful in EEG-based research because they can handle high-dimensional inputs and model patterns that are difficult to capture using basic linear models. In this paper, we use an SVM classifier using the Radial Basis Function (RBF) as its kernel. This is because EEG data frequently exhibits nonlinear relationships that cannot be adequately represented by a straight-line decision boundary. The RBF kernel assists the classifier in detecting these tiny differences by mapping the data into a space where the classes are more separable. This made it an ideal option for our emotion-classification challenge, as brain-signal characteristics tend to exhibit complicated and sometimes interlocking patterns. 
Working with SVMs has practical challenges, including reduced performance as datasets grow, particularly when nonlinear kernels are used. 
\subsubsection{Random Forest} \label{Sec:RandomForest}
This method uses ensemble learning to improve generalization in comparison to single decision trees. The ensemble learning involves training multiple decision trees on various data and feature sets. The classification choice is made by a majority vote from all trees in the ensemble~\cite{breiman2001random}. 
In Random Forest (RF) classification, the final predicted label $\hat{y}$ for an input sample $\mathbf{x}$ is obtained by aggregating the outputs of an ensemble of $T$ decision trees~\eqref{eq:RF}. Each tree $h_t(\mathbf{x}), t = 1, \ldots, T,$ independently produces a class prediction based on a bootstrap sample of the training data and a random subset of features. The RF classifier assigns $\mathbf{x}$ to the class that receives the majority vote among all trees, which can be expressed as
\begin{equation}
    \label{eq:RF}
    \hat{y} = \operatorname{mode}\big(h_1(\mathbf{x}), h_2(\mathbf{x}), \ldots, h_T(\mathbf{x})\big).
\end{equation}
This ensemble voting mechanism improves classification robustness and generalization by reducing variance and mitigating overfitting compared to individual decision trees.

The RF model offers various advantages for EEG-based emotion classification. It trains each decision tree on a separate section of the dataset, and each split within the tree uses a randomly selected collection of characteristics. Randomness ensures that each tree learns unique patterns, resulting in a more stable and durable model. 
EEG studies can determine which electrode positions, frequency ranges, or characteristics are most effective in distinguishing emotional states. The RF models excel at handling complex datasets by capturing non-linear correlations and feature interactions that linear models typically overlook. Even with a wide range of data formats and high variability in feature values, performance remains consistent. The RF models, unlike SVM models, are less sensitive to feature scaling, making preprocessing easier and increasing workflow flexibility. However, this strategy has its limitations. The model's complexity grows in proportion to the number of features. Training time may be longer depending on the number of trees. The RFs may struggle with highly correlated features, which is a common problem in EEG datasets where adjacent electrodes record similar brain activity.

\subsection{Model Training and Evaluation Procedure}\label{Sec:ModelTraining}
All three ML models are implemented in Python using its Scikit-learn library. For a fair comparison, the same preprocessing, cleaning, normalization, and feature extraction are applied before training the models. The final performance is evaluated with the F1-score and accuracy parameter.

To ensure fairness, all three models are compared using the same train-test split. To ensure consistency in the split over multiple tests, a fixed random seed is used. This minimized variation in results caused by data splitting. The sensitivity of the SVM classifier to feature scale necessitated feature normalization. To improve performance using the RF model, SVM hyperparameters are tuned. After training, the models are tested using accuracy scores, confusion matrices, and other metrics to assess their ability to accurately classify the three emotional states. Comparing the findings helps us to identify areas where each model excels and where it fails. This offers a better understanding of how different algorithms handle EEG signal complexity and noise.

\subsection{Dataset}\label{Sec:Dataset}
\begin{figure}
    \centering
    \includegraphics[width=0.95\linewidth]{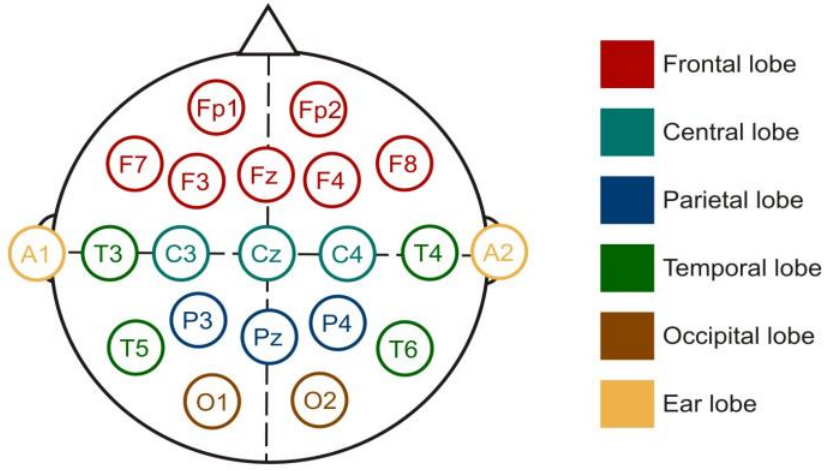}
    \caption{EEG electrode positions according to the international 10-20 system standard~\cite{murtazina2020ontology}.}
    \label{fig:EEG_Positions}
    \vspace{-.4cm}
\end{figure}
 
\begin{figure}
\centering
\includegraphics[width=0.405\textwidth]{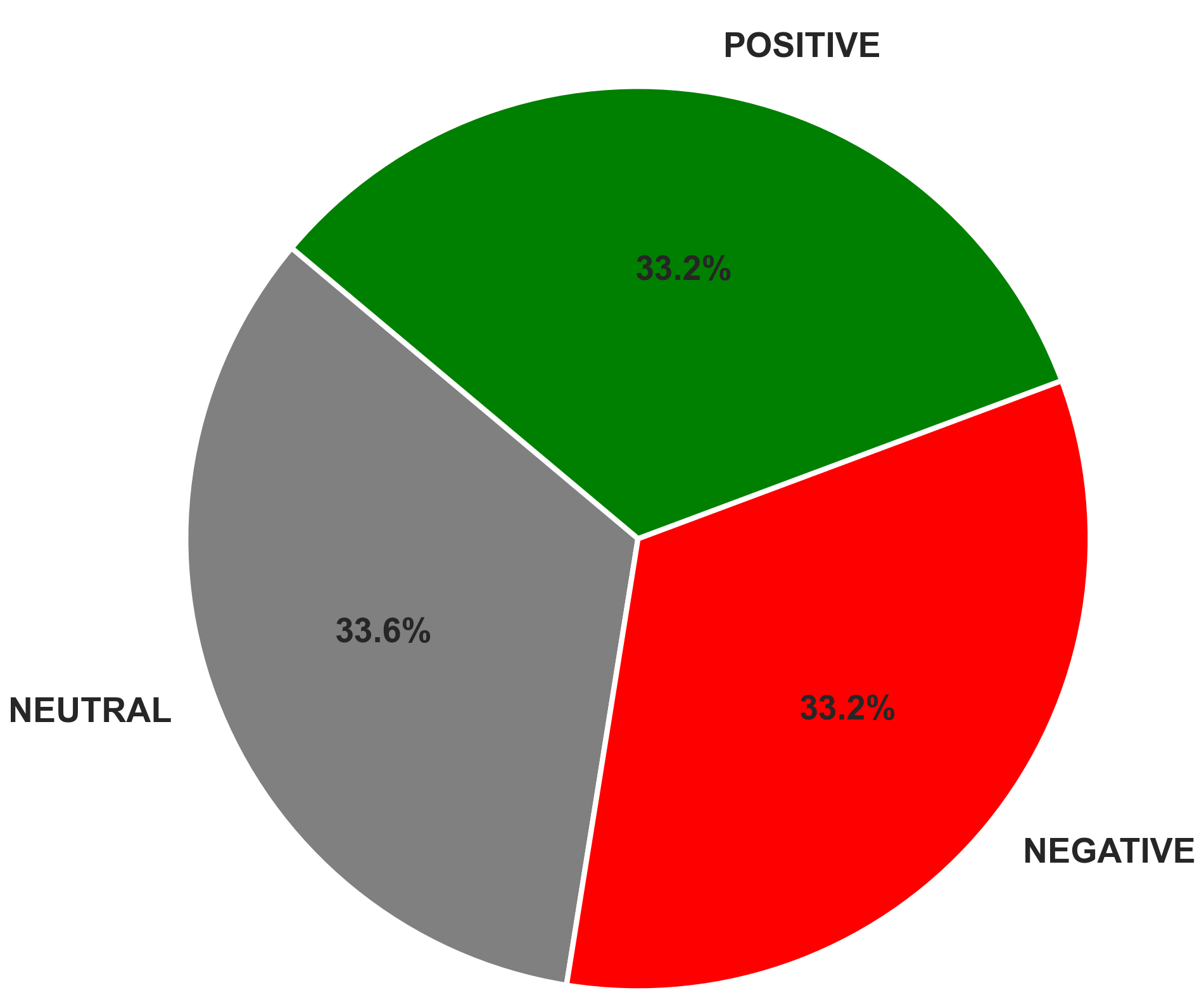}
    \caption{The distribution of emotions in the dataset.}
    \vspace{-.4cm}
\label{fig:emotion_distribution}
\end{figure}

The dataset used for this study contains EEG, also known as brainwave recordings, that have been processed through a statistical feature extraction technique based on a preprocessing strategy described in~\cite{bird2019mental}. Our research uses EEG data that was recorded using a Muse EEG headband, a consumer-grade, multi-channel EEG device~\cite{zhang2019effect}. In our work, we use EEG electrode names and positions as per the international 10--20 system standard (see Fig.~\ref{fig:EEG_Positions}).\footnote{The electrode names start with the first letter of the Latin name of the area where the electrode is placed, followed by a number indicating the electrode's side and location within that area. The areas used are prefrontal (Fp), frontal (F), temporal (T), parietal (P), occipital (O), central (C), midline (Z), and auricular (A). EEG focuses on delta, theta, alpha, beta, and gamma frequency ranges. Alpha and beta waves are the most common in the EEG of a healthy adult, whether at rest or awake.
}
For each of the three emotional states (Positive, Neutral, and Negative) elicited by particular video clips, EEG data was obtained from two volunteers (one male, one female, ages 20 to 22). The dataset consists of 3,24,000 data points that were resampled to 150 Hz from brain waves. Additionally, neutral brainwave data was gathered as a third emotional state, Neutral, that reflected the patients' emotional states during rest. Four electrode placements (TP9, AF7, AF8, and TP10), in accordance with the international 10-20 EEG system, captured the EEG data, which was then processed to produce a dataset of statistical features extracted using a one-second sliding window, starting at $t=0$ and $t=0.5$. This process produced a well-balanced dataset across all three emotional states (see Fig.~\ref{fig:emotion_distribution}) that, besides EEG features, included emotion labels, namely Positive, Neutral, and Negative.



Next, participants selected a priori are exposed to a set of carefully chosen audiovisual stimuli in the form of video clips to elicit the targeted emotional states. The emotional categories and the corresponding stimuli are listed in Table~\ref{tab:emotions}.

\begin{table}
    \centering
    \caption{The emotional categories and the corresponding stimuli.}
    \begin{tabular}{|c|l|l|}
    \hline
        Emotion & Movie Video Clip & Source/Description\\ \hline
        Negative & Marley and Me & Death Scene \\ & &(Twentieth Century Fox)\\ \hline
        Negative & Up & Opening Death Scene \\ & & (Walt Disney Pictures)\\ \hline
        Negative & My Girl & Funeral Scene \\ & &(Imagine Entertainment)\\ \hline
        Positive & La La Land & Opening Musical Number \\ & & (Summit Entertainment)\\ \hline
        Positive & Slow life & Nature Timelapse \\ & & (BioQuest Studios)\\ \hline
        Positive & Funny Dogs & Compilation of Funny Dog Clips \\ & & (MashupZone)\\ \hline
        Neutral & All movies & Resting-state \\ \hline
    \end{tabular}
    \label{tab:emotions}
\end{table}

\section{Results and Discussion}\label{Sec:Results}
In this section we provide the results, which are generated through coding in the Python language. The list of libraries used is as follows: 
\begin{itemize}
    \item Pandas and Numpy for data handling
    \item Matplotlib and Seaborn for visualization
    \item Scikit-learn for ML models
    \item Tensorflow for neural network implementation (for extended comparison)
    \item Scipy for signal processing and t-tests
\end{itemize}

\begin{figure}
\centering
\includegraphics[width=0.485\textwidth]{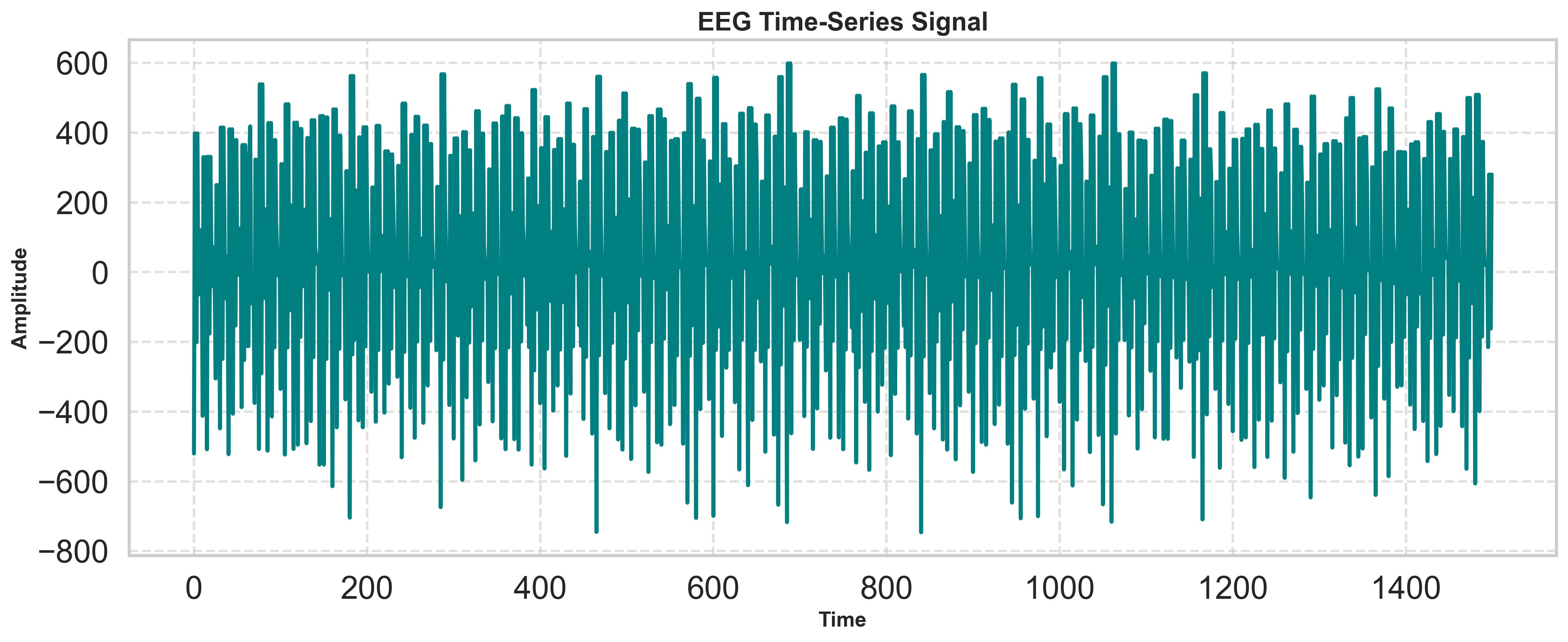}
\caption{This plot shows EEG time series of the cleaned data.}
\label{fig:eeg_time_series}
\vspace{-.4cm}
\end{figure}

\begin{figure}
\centering
\includegraphics[width=0.485\textwidth]{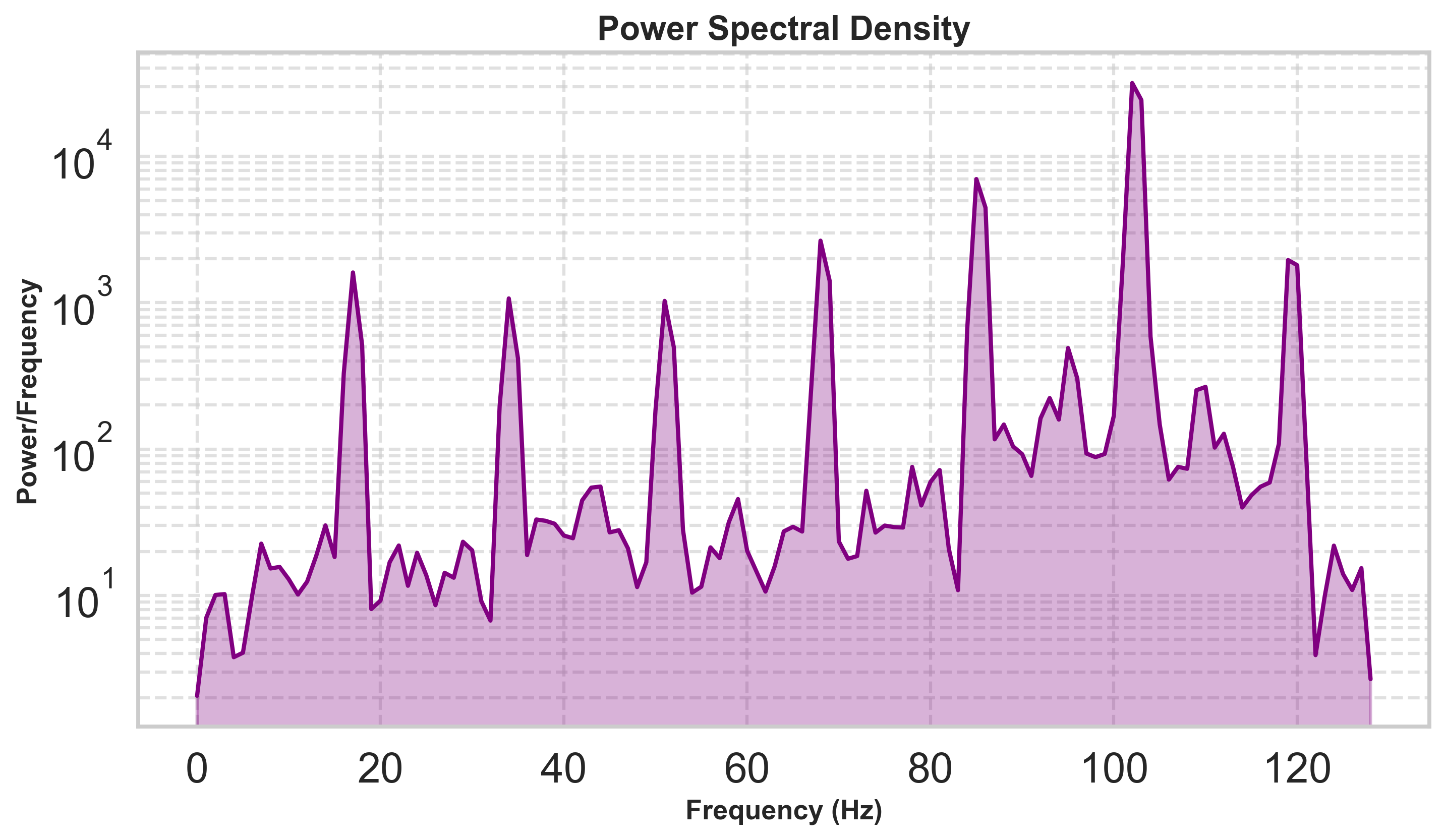}
\caption{This plot shows the power spectral density (PSD) of the EEG signals.}
\label{fig:power_spectral_density}
\vspace{-.4cm}
\end{figure}

To observe the behavior of raw EEG signals across different channels and emotional states over time, EEG signals from 19 channels per subject are loaded and processed. Bandpass filters are applied to remove noise from the signals, and the cleaned data are then visualized (as shown in Fig.~\ref{fig:eeg_time_series}) to analyze signal variations in response to different emotional stimuli. 
Next, to examine how the power of EEG signals is distributed across different frequency bands, including Delta, Theta, Alpha, Beta, and Gamma, the power spectral density (PSD) analysis is performed. To accomplish this, the Fast Fourier Transform (FFT) is applied to the EEG signals, PSD is computed (as shown in Fig.~\ref{fig:power_spectral_density}) for each electrode and time window, and frequency-specific energy patterns (features) are extracted for further analysis. From Fig.~\ref{fig:power_spectral_density}, we can observe that emotional states are more prominent in Beta (13–-30 Hz)  and Gamma ($>$30 Hz) regions.



\begin{figure}
    \centering
    \includegraphics[width=0.995\linewidth]{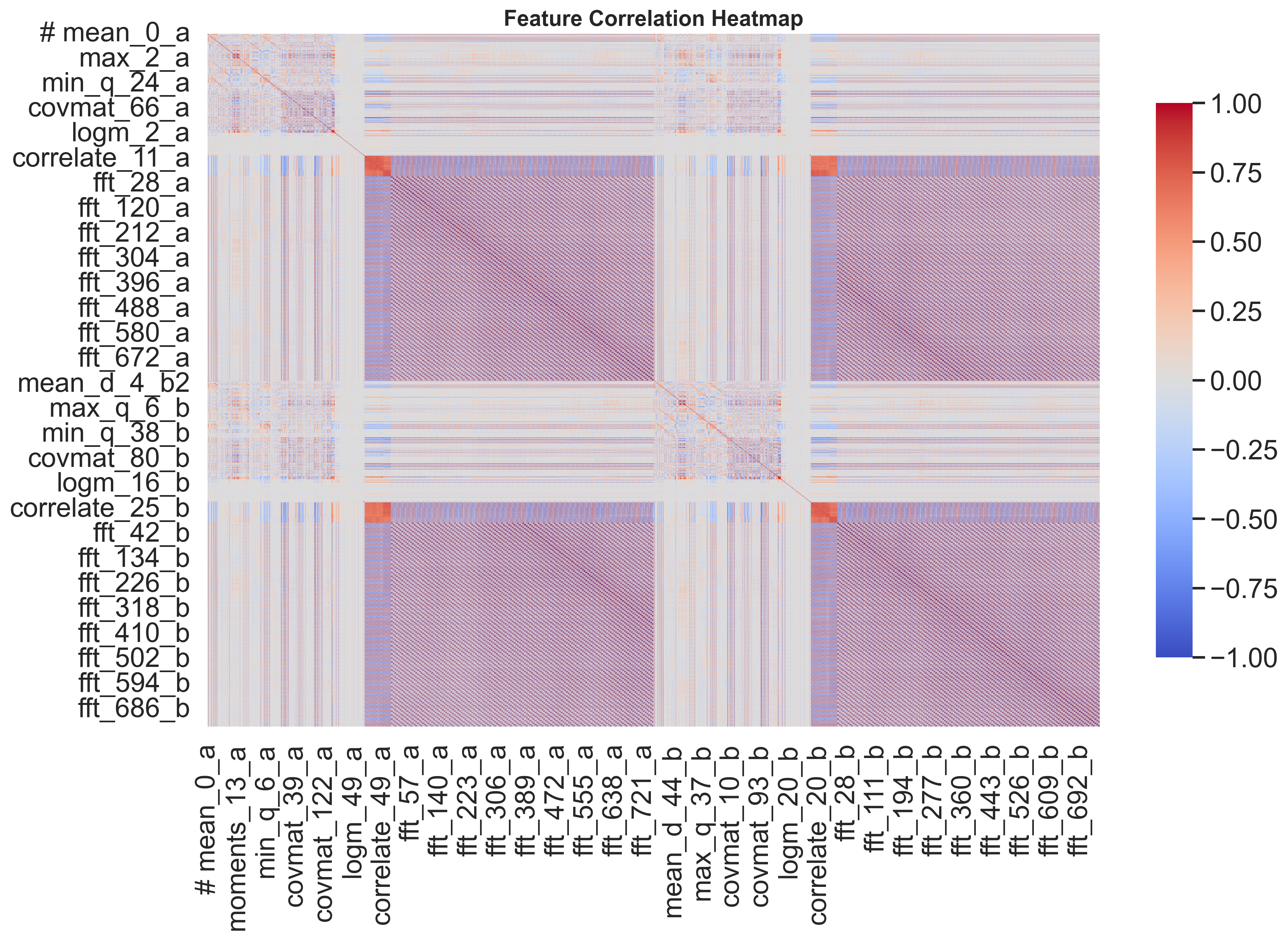}
    \caption{This plot shows the correlation heatmap between different features. 
    }
    \label{fig:correlation_heatmap}
    \vspace{-.2cm}
\end{figure}
\begin{figure}
    \centering
    \includegraphics[width=0.95\linewidth]{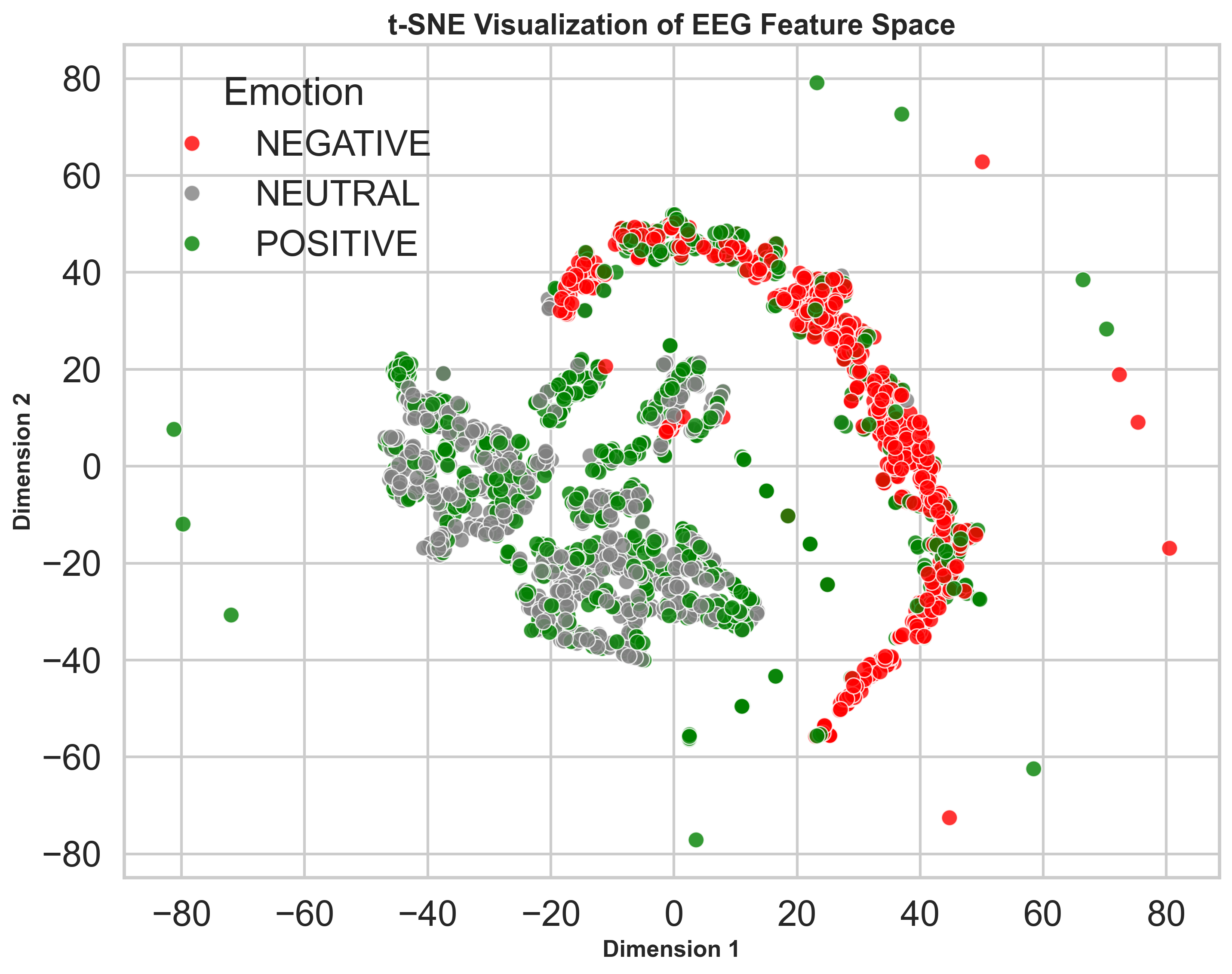}
    \caption{This plot shows the t-SNE visualization to visualize the separability of different emotions in a 2D space.}
    \label{fig:tsne_visualization}
    \vspace{-.4cm}
\end{figure}

Next, we perform the correlation analysis, which is aimed at understanding the relationships among the extracted EEG features and at identifying potential redundancy. This is achieved by computing correlation coefficients (range $-1$ to $+1$) between the features and visualizing the results using a heatmap as shown in Fig.~\ref{fig:correlation_heatmap}.
Next, the t-distributed Stochastic Neighbor Embedding (t-SNE) is used to visualize the separability of EEG features across different emotion classes in a two-dimensional space. The method involved applying t-SNE for dimensionality reduction and plotting the resulting feature clusters, which revealed clear separation for some emotions, in particular between Neutral and Negative emotions,  while showing overlap for the remaining, as shown in Fig.~\ref{fig:tsne_visualization}.

\begin{figure}
\centering
\includegraphics[width=0.485\textwidth]{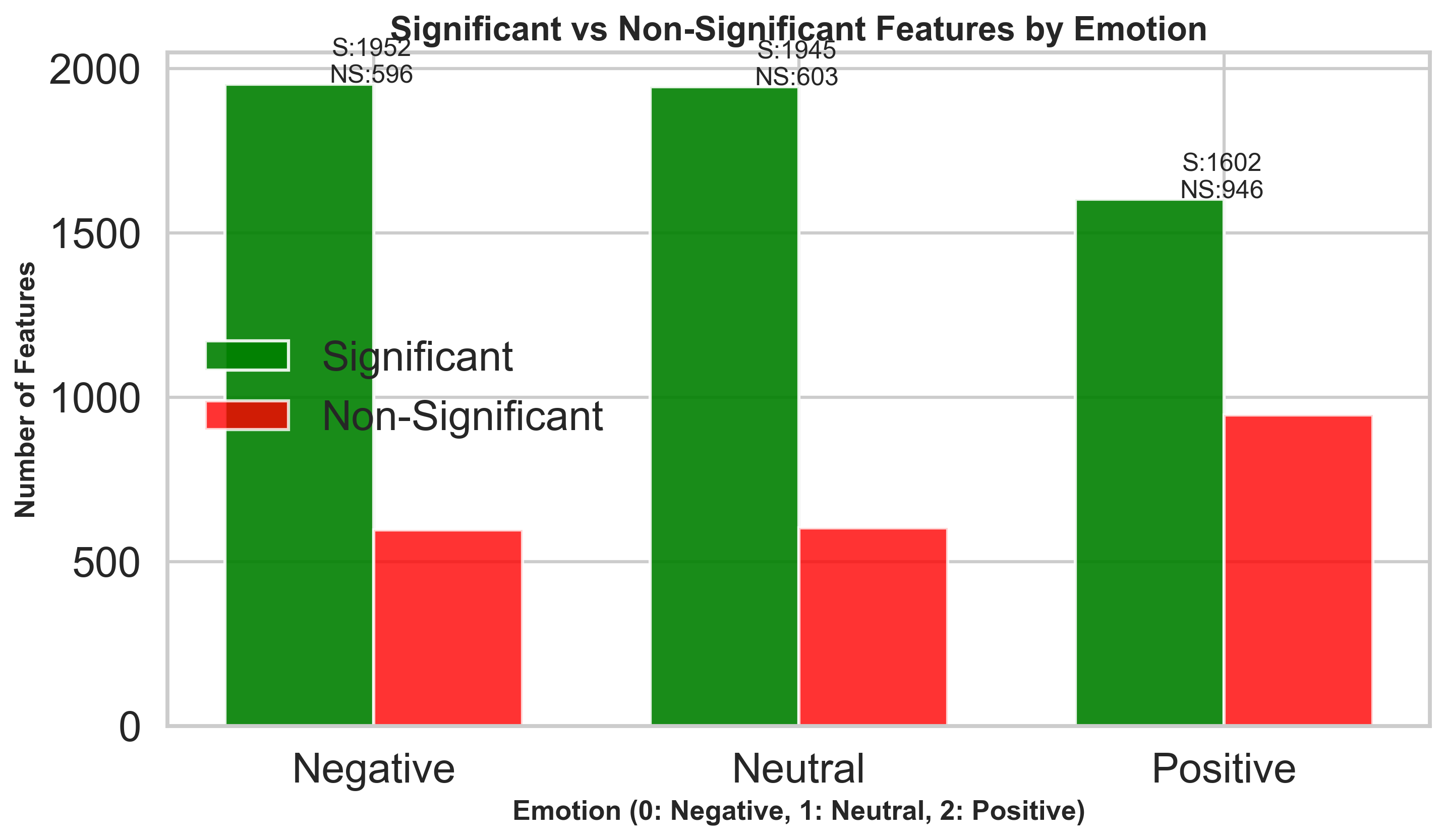}
\caption{This plot shows the comparison between significant and non-significant features per emotion.}
\label{fig:significant_features_by_emotion}
\vspace{-.4cm}
\end{figure}

Statistical testing is conducted to identify EEG features that are statistically significant in distinguishing between emotions. This involved applying t-tests between emotion classes (0: Negative, 1: Neutral, 2: Positive), counting the number of statistically significant features ($S$) versus non-significant features ($NS$), and plotting the results as shown in Fig.~\ref{fig:significant_features_by_emotion}. From the figure it is clear that the significant features play an important role in emotion classification. 
\begin{figure}
\centering
\includegraphics[width=0.375\textwidth]{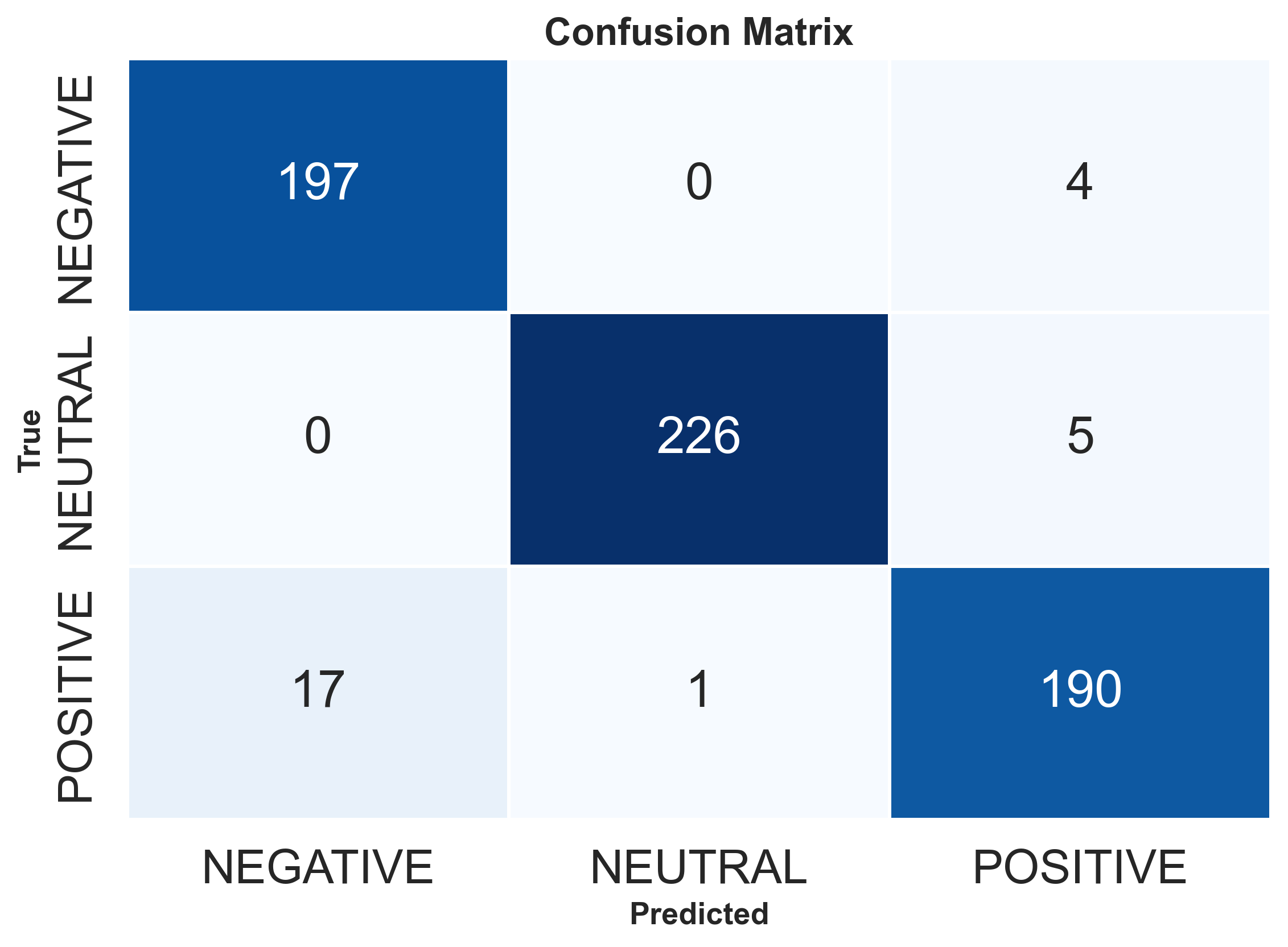}
\caption{This plot shows the confusion matrix of the LR model.}
\label{fig:confusion_matrix}
\vspace{-.2cm}
\end{figure}
Next, the confusion matrix is used to evaluate the performance of the emotion classification models in correctly identifying Negative, Neutral, and Positive emotional states from EEG signals. It is generated by comparing the predicted emotion labels with the true labels, thereby illustrating the accuracy of the model as well as misclassification patterns across the three emotion classes. In Fig.~\ref{fig:confusion_matrix}, as an example, we provide the confusion matrix of the LR model.


\begin{table}
\vspace{-.2cm}
\caption{ML Model Comparison Table}
    \centering
    \resizebox{0.805\columnwidth}{!}{%
    \begin{tabular}{|l|c|c|}
    \hline
        Models & Accuracy (\%) & F1-score  \\ \hline
        Logistic regression & 95.6 & 0.956 \\ \hline
        SVM & 93.9 & 0.938 \\ \hline
        Random Forest & 97.5 & 0.975 \\ \hline
    \end{tabular}%
    }
    \label{Tab:Accuracy_Compare}
    \vspace{-.2cm}
\end{table}

Now, we apply ML models to the preprocessed data for classification among the three emotion categories, viz., Positive, Negative, and Neutral. The performance of the three ML models, namely LR, SVM, and RF, in terms of classification accuracy and F1-score is determined and displayed the same in Table~\ref{Tab:Accuracy_Compare}. From the results, we can infer that the RF model achieved the best accuracy and F1-score, indicating strong classification ability for EEG-based emotions. 
The reason for the RF model to perform the best is probably because it forms flexible and non-linear decision boundaries fitting the complex and high-dimensional nature of the EEG features. The LR model yielded quite consistent results, although it showed slight signs of overfitting on the training set. The SVM model ran efficiently and is quick to train, but its accuracy is lower compared to the other two models.

Lastly, we also compare the performance of our best-performing model, which is the RF model, with the models developed in the literature on the similar data set. The results are shown in Table~\ref{Tab:Others_Compare}. From the results, we can infer that the RF model outperforms other models specified in Table~\ref{Tab:Others_Compare}.

\begin{table}
\vspace{-.2cm}
\caption{Performance comparison of Random Forest with the other state-of-the-art models}
    \centering
    \resizebox{0.805\columnwidth}{!}{%
    \begin{tabular}{|l|c|}
    \hline
        Models & Accuracy (\%) \\ \hline
        Deep Belief Network~\cite{zheng2014eeg} & 87.6 \\ \hline
        Common Spatial Patterns~\cite{li2009emotion} & 93.5  \\ \hline
        InfoGain MLP~\cite{bird2019mental} & 94.5 \\ \hline
        Fisher’s Discriminant~\cite{oude2006eeg} & 94.9  \\ \hline
        Conformer model~\cite{song2022eeg} & 95.3  \\ \hline 
        Random Forest & 97.5  \\ \hline
    \end{tabular}%
    }
    \label{Tab:Others_Compare}
    \vspace{-.2cm}
\end{table}


\section{Conclusions and Future Work}\label{Sec:Conclusions}
This study demonstrated how well machine learning models can categorize EEG signals into different emotional states. The RF model performed the best among the models that are compared, followed by the LR and the SVM models.
Future plans include capturing temporal EEG dynamics using deep-learning architectures like CNN and LSTM. To improve generalization, use bigger datasets and investigate hybrid models that combine neural and statistical methods.

\bibliographystyle{ieeetr}
\bibliography{references.bib}
\end{document}